\documentclass[letterpaper, 10 pt, conference]{ieeeconf}  

\IEEEoverridecommandlockouts                              

\usepackage{amsmath}
\usepackage{url}
\usepackage{booktabs}
\usepackage[dvipsnames]{xcolor}
\usepackage{tcolorbox} 
\usepackage{colortbl} 

\usepackage{algorithm}
\usepackage{algorithmic}

\usepackage{subcaption} 
\usepackage{cleveref}

\usepackage{tikz}
\usepackage{pgfplots}
\pgfplotsset{compat=1.18}

\usepgfplotslibrary{groupplots,polar,statistics}

\pgfplotsset{
  customlegend/.style={
    legend image code/.code={
      \draw[##1,fill=##1] (0cm,-0.1cm) rectangle (0.3cm,0.2cm);
    }
  }
}

\usepackage{CJKutf8}

\title{\LARGE \bf
J-ORA: A Framework and Multimodal Dataset for Japanese Object Identification, Reference, Action Prediction in Robot Perception
}

\author{Jesse Atuhurra$^{1,2}$, Hidetaka Kamigaito$^{1}$, Taro Watanabe$^{1}$ and Koichiro Yoshino$^{1,2}$
\thanks{$^{1}$Division of Information Science, NAIST, Nara, Japan
        }%
\thanks{$^{2}$Guardian Robot Project, RIKEN,
        Kyoto, Japan
        }%
\thanks{The corresponding author is Jesse Atuhurra. The email is: {\tt\small atuhurra.jesse.ag2@naist.ac.jp} }
}

\begin{document}

\maketitle
\thispagestyle{empty}
\pagestyle{empty}

\begin{abstract}
We introduce J-ORA, a novel multimodal dataset that bridges the gap in robot perception by providing detailed object attribute annotations within Japanese human-robot dialogue scenarios. 
J-ORA is designed to support three critical perception tasks, object identification, reference resolution, and next-action prediction, by leveraging a comprehensive template of attributes (e.g., category, color, shape, size, material, and spatial relations). 
Extensive evaluations with both proprietary and open-source Vision Language Models (VLMs) reveal that incorporating detailed object attributes substantially improves multimodal perception performance compared to without object attributes. 
Despite the improvement, we find that there still exists a gap between proprietary and open-source VLMs.
In addition, our analysis of object affordances demonstrates varying abilities in understanding object functionality and contextual relationships across different VLMs. 
These findings underscore the importance of rich, context-sensitive attribute annotations in advancing robot perception in dynamic environments.
See project page at https://jatuhurrra.github.io/J-ORA/. 
\end{abstract}

\section{INTRODUCTION}
Due to advances in vision-and-language model (VLM) research based on deep learning technologies, robots can now do various tasks to help us in our living spaces. Real-world robots recognize the objects around them, understand their properties and the user's relationship with them, and plan the required actions \cite{patterson2016coco, 5206772, 6909715, 5206594, krishna2016visualgenomeconnectinglanguage}.

Many existing VLMs have been realized through the object recognition task with object labels or contrastive learning with captions attached to situations. However, these types of training are insufficient for generalizing the robot's performance. When humans encounter an unknown object, we try to guess its properties from its shape, color, size, position, etc., and use them. If a robot can appropriately use this kind of information, it will have a better understanding of its cognition and the situation in which it is placed. It will also be able to improve the accuracy of its action planning using this information.

Object attributes have been used for first-person scene understanding for robots. Semantic attributes such as shape, color, and geographical information \cite{5206594} are hand-crafted according to the situation and task \cite{patterson2016coco}. We must handle and refine the information about these attributes from the perspective of robot action planning. In this study, we define three types of tasks to verify the degree to which attribute information contributes to robot perception. We also introduce an attribute information template for defining the attribute information required for these robot tasks, and annotate it for the user-robot interaction situation in an actual living space.
\begin{figure}[t!]
    \centering
    \includegraphics[width=0.995\linewidth]{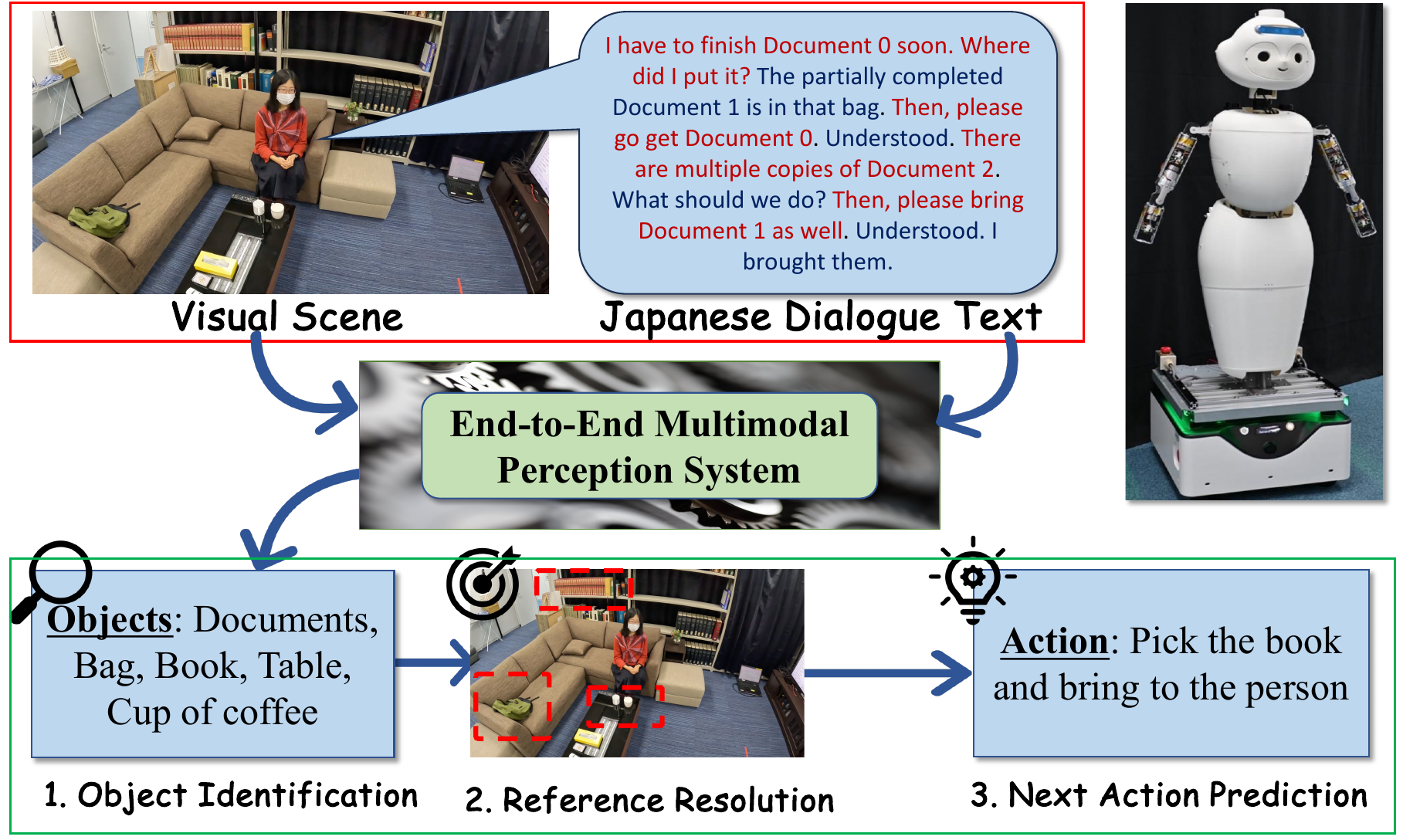}
    \caption{We investigate the contribution of object attributes towards the robot's perception, and introduce three perception tasks to facilitate the analysis. }
    \label{fig:EndtoEndPerception}
\end{figure}

We define three tasks: object recognition, reference resolution, and action prediction, corresponding to the three stages of robot recognition, understanding, and planning. In object recognition, the robot recognizes objects, their positions, and their attributes in images obtained from a first-person perspective. This is essential for planning, manipulation, and safe navigation. In reference resolution, the robot maps object rectangles to references made during conversations between the robot and the user.
By achieving this task, it will be possible to perform user assistance while communicating appropriately in an environment where the robot lives with the user. In action prediction, the robot infers the appropriate action to take next based on the context of the dialogue, the information it can currently observe, and its understanding.

We redefine the definition of attribute information for the objects the robot should handle and create templates for these attribute annotations to improve these tasks. To acquire a diversity of attributes, we combine the \textit{general attributes} of objects (such as color, shape, size) with the \textit{specific attributes} of electronic devices (such as state, type, interface elements), inside decor (such as style, artistic elements), personal items (such as portability, personal value), and persons (such as posture, facial expression, interaction with objects). We also include the \textit{spatial} and \textit{functional} relation between objects and the human subjects in the visual scene. It also includes the spatial and functional relationships between objects and human subjects in visual scenes in living spaces. Based on these templates, we extend a multimodal coreference analysis dataset (J-Cre3)~\cite{ueda-etal-2024-j} recorded as actual user-robot dialogue.

This research contributes the following: (i) We construct an annotation schema for attribute information that assumes a situation where a robot assists a user interactively in an indoor environment and then applies it to an actual dialogue dataset. This annotation was carried out on a Japanese dialogue dataset, but there are no large-scale datasets of this kind of attribute information for Japanese. (ii) We measured the contribution of this data with defined attributes to the existing VLM capabilities on our three tasks: object recognition, reference resolution, and action prediction. We used proprietary models such as GPT-4o, open models developed in English, and open models developed in Japanese. We also implemented joint optimization of these three tasks in fine-tuning of VLMs, to realize end-to-end multimodal perception model for robots.

Our experiments revealed that introducing object attributes improves object recognition and reference resolution through a deep understanding of visual context and object grounding. In addition, there was a significant difference between the closed and open models, highlighting the challenges of end-to-end multimodal perception in Japanese dialogue. Furthermore, the analysis of object affordances revealed differences in the ability to understand objects' functionality and contextual relationships between different visual language models.

\begin{figure*}[t!]
\centering
\includegraphics[width=0.895\linewidth]{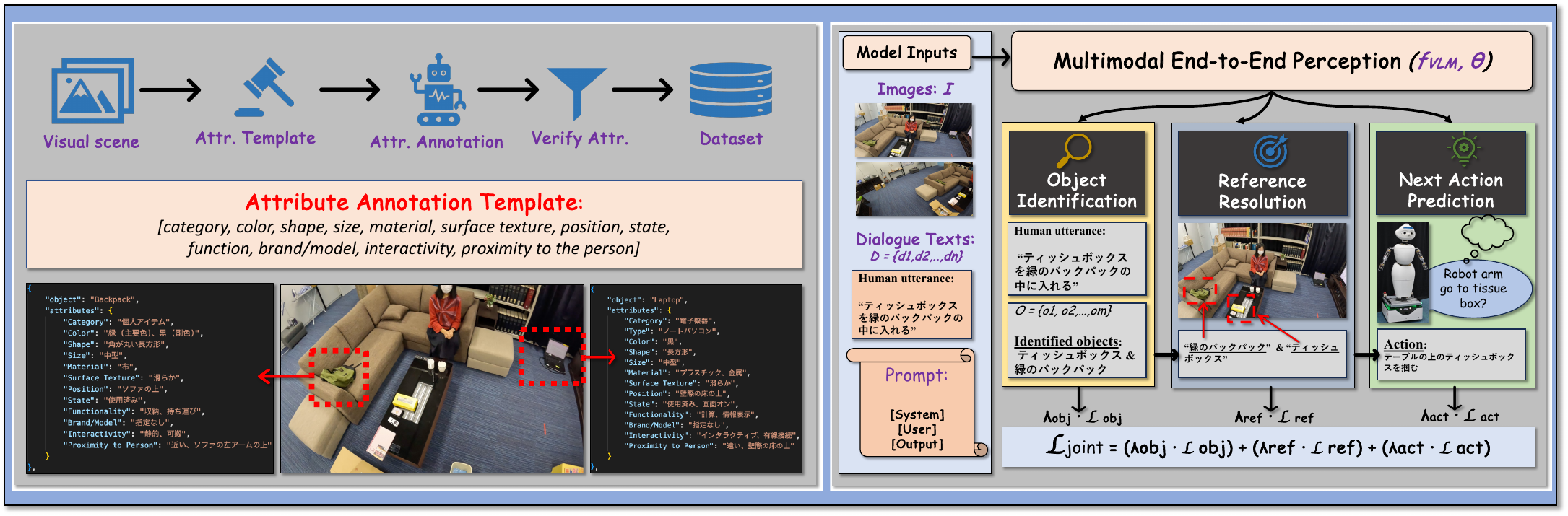}
\caption{ \textbf{Left:} We develop a \textit{template of object attributes} and use it to annotate objects in Japanese, creating J-ORA. A highlighted instance from J-ORA at the bottom shows two objects (backpack and laptop). The scene comprises the robot's egocentric view. \textbf{Right:} Our \textit{multimodal end-to-end perception system}, based on a VLM, consists of three tasks that are performed simultaneously. 
System inputs include images/videos, dialogue utterances, and prompts.
The human utterance above translates into \textit{`Place the tissue box in the green backpack.'} }
\label{fig:PaperMethodObjAttr}
\end{figure*}
\section{RELATED WORK}
\subsection{Object Attribute Datasets}
Researchers have previously curated datasets that contain deliberate annotations for \textit{object attributes}. 
Examples of such datasets include \cite{krishna2016visualgenomeconnectinglanguage, patterson2016coco, 5206772, 5206594, WahCUB_200_2011, 10.1007/s11263-013-0695-z, liuLQWTcvpr16DeepFashion, openimages, Pham_2021_CVPR}.
Object detection, recognition, and scene understanding are essential to perception. 
\cite{krishna2016visualgenomeconnectinglanguage}, released about a decade ago, introduced annotations for scene understanding, visual question answering, and relationships between objects and their attributes. The data contains over 100K images and attribute annotations are of type \textit{emotion, pose, color, size, material, state, texture, pattern, etc.,}, so each image has an average of 18 attributes.
While precious, existing datasets lack a perspective from the robot inside scenes containing dynamically changing object states and attributes.
\subsection{Multimodal Dialogue Datasets}
The datasets contain dialogue data useful for training a robot perception system. Below, we categorize them based on how the data is collected.
\textit{(i) Human-Robot Dialogues.} Acquiring actual interactions between humans and robots is expensive but immensely valuable in capturing the nuances existent in human language. Datasets include \cite{9223340, irfan2024humanrobotinteractionconversationaluser, 10182264, ALFRED20, deruyttere2019talk2car}. 
\textit{(ii) Human-Human Dialogues.} In the absence of a robot, researchers still manage to collect human-to-human role-playing dialogues that mimic human-and-robot interactions. Datasets include \cite{lizi2021mmconv, Moon2019opendialkg}.
\textit{(iii) Simulated Dialogues.} Photo-realistic environments and simulators have been used to create multimodal conversational data. \cite{kottur-etal-2021-simmc} and \cite{TEACh22} comprise pairs of dialogue utterances and visual data collected using a simulator. 
\textit{(iv) Additional Datasets.} 
They contain speech/textual and image/video data, and include \cite{alamri2019audiovisual, VQA, visdial, kebe2021a, kottur-etal-2019-clevr}. 
The Japanese-specific datasets include \cite{inadumi-etal-2024-gaze-grounded, C18-1163}. 
\textit{However, none of the existing datasets contain the rich attributes needed for the Japanese end-to-end perception task considered in this paper.}
\subsection{Multimodal Perception}
Integrating multimodal inputs, such as vision and language, is essential for robots to perceive scenes and plan actions effectively in human-robot interaction scenarios. Previous works have explored end-to-end approaches to guide robot behavior in such contexts. For instance, \cite{10.1145/3613905.3651029} presented a system that combines multiple sensory modalities to control robot actions during interactions. Similarly, \cite{tanneberg2024helphelpllmbasedattentive} leveraged GPT-4 to process scene and dialogue information for situation understanding and action generation, though their approach is limited to English and does not explicitly utilize detailed object attribute information.
\subsection{Robot Action Planning}
Research demonstrates that object attributes, such as material and shape, are vital for robots to tailor actions and deduce possible interactions, enhancing task planning \cite{9981578}, \cite{10801408}, and affordance learning \cite{Bahl_2023_CVPR}, \cite{6225042}, \cite{10656869}.
Effective action planning in ever-changing scenes relies on systems that monitor object state shifts, with datasets like J-ORA providing essential annotations of these evolving attributes.
\textit{J-ORA improves upon previous datasets by offering attribute annotations within dynamic Japanese human-robot dialogue contexts, boosting multimodal perception for sophisticated action planning}.
\section{PRELIMINARIES FOR END-TO-END MULTIMODAL PERCEPTION}
We propose a new approach that leverages the latest developments in VLMs to construct a perception system that includes \textbf{O}bject Identification, Object 
\\ \textbf{R}eference, and \textbf{A}ction Prediction, 
to perform multiple tasks end-to-end.  
As shown in Figure \ref{fig:PaperMethodObjAttr}, the system is based on VLMs, and the inputs include the (i) visual scene, (ii) dialogue texts, and (iii) textual prompts. 
The parameters inside the VLM, $\theta$, are updated during fine-tuning to perform three tasks simultaneously. 
\\ \textbf{Object Identification:}
Let \(\mathcal{D} = \{d_1, d_2, \dots, d_n\}\) represent a sequence of dialogue utterances, where each \(d_i\) is a text input. The goal is to identify a set of objects \(\mathcal{O} = \{o_1, o_2, \dots, o_m\}\) from a predefined object vocabulary \(\mathcal{V}_o\), and each object is annotated with attributes. 
The identification task is framed as a sequence labeling problem, where a function \(f_{\text{obj}}\) maps each dialogue utterance \(d_i\) to a subset of \(\mathcal{V}_o\):
\[
f_{\text{obj}}(d_i) = \{o_j \mid o_j \in \mathcal{V}_o, p(o_j \mid d_i) > \tau_{\text{obj}}\}
\]
Here, \(p(o_j \mid d_i)\) is the probability that object \(o_j\) is mentioned in \(d_i\), and \(\tau_{\text{obj}}\) is a threshold for object identification.
\\ \textbf{Reference Resolution:} 
Given an identified object \(o_j\) from the previous task and a visual context \(\mathcal{I}\) (the image or set of images representing the robot's surroundings), the task is to map the textual reference of the object to its corresponding visual representation. Define a function \(f_{\text{ref}}\) that maps the object \(o_j\) and visual context \(\mathcal{I}\) to a specific region \(\mathcal{R} \subset \mathcal{I}\) in the image determined by the maximum likelihood estimate:
\[
\mathcal{R}_{o_j} = \arg\max_{\mathcal{R} \subset \mathcal{I}} p(\mathcal{R} \mid o_j, \mathcal{I}) \quad \text{such that} \quad f_{\text{ref}}(o_j, \mathcal{I}) = \mathcal{R}_{o_j}
\]
where, \(\mathcal{R}_{o_j}\) represents the region in the image \(\mathcal{I}\) that corresponds to the object \(o_j\).
\\ \textbf{Next Action Prediction:}
The task is to predict the next action \(a_{t+1} \in \mathcal{A}\) (where \(\mathcal{A}\) is a set of possible actions) based on the identified objects \(\mathcal{O}_{1:t}\), the visual context \(\mathcal{I}\), and the sequence of previous actions \(\mathcal{A}_{1:t}\). Let \(f_{\text{act}}\) denote the action prediction function:
\[
f_{\text{act}}(\mathcal{O}_{1:t}, \mathcal{I}, \mathcal{A}_{1:t}) = \arg\max_{a \in \mathcal{A}} p(a \mid \mathcal{O}_{1:t}, \mathcal{I}, \mathcal{A}_{1:t})
\]
Here, \(p(a \mid \mathcal{O}_{1:t}, \mathcal{I}, \mathcal{A}_{1:t})\) represents the conditional probability of action \(a\) given the identified objects, visual context, and action history.
\\ \textbf{Joint Optimization:}
We propose a joint objective function that integrates the three tasks to optimize the end-to-end system. 
Let \(\mathcal{L}_{\text{obj}}\), \(\mathcal{L}_{\text{ref}}\), and \(\mathcal{L}_{\text{act}}\) be the loss functions for the object identification, text-to-image reference, and next action prediction tasks, respectively. The joint objective function is expressed as:
\[
\mathcal{L}_{\text{joint}} = \lambda_{\text{obj}} \cdot \mathcal{L}_{\text{obj}} + \lambda_{\text{ref}} \cdot \mathcal{L}_{\text{ref}} + \lambda_{\text{act}} \cdot \mathcal{L}_{\text{act}}
\]
where \(\lambda_{\text{obj}}\), \(\lambda_{\text{ref}}\), and \(\lambda_{\text{act}}\) are the weighting factors for each task's loss.

The optimization goal is:
\[
\theta^* = \arg\min_{\theta} \mathcal{L}_{\text{joint}}(f_{\text{VLM}}(\theta; \text{prompts}), \mathcal{D}, \mathcal{I}, \mathcal{A})
\]
where \(\theta\) represents VLM parameters involved in three tasks. Prompts guide VLM to perform tasks effectively.

\section{J-ORA DATASET}
\label{sec:datacollection}
\begin{figure*}[t!]
    \centering
    \includegraphics[width=0.795\linewidth]{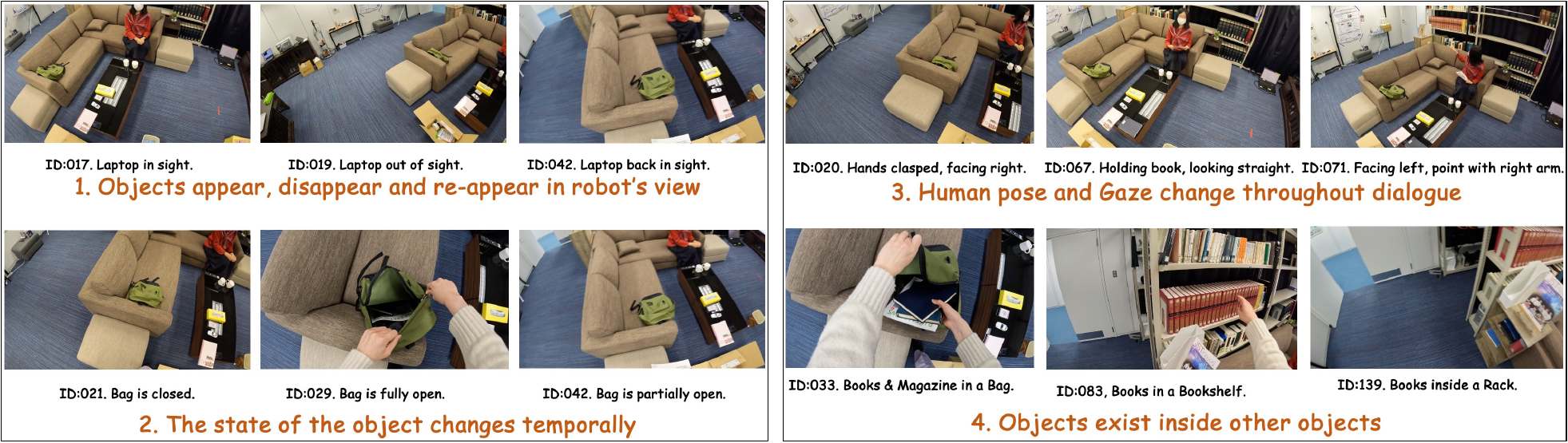}
    \caption{Changes that occur in the scene during dialogue. We have shown image IDs for reference to image sequences.}
    \label{fig:PaperObjectChanges}
\end{figure*}
We aim to develop a dataset sufficient to train and evaluate end-to-end multimodal perception systems.
The system inputs consist of (i) dialogue texts extracted from audio transcripts of the dialogue, (ii) the images extracted from videos of the dialogue scene, and (iii) the prompts needed to elicit appropriate responses from the VLM. 
The output includes the (i) objects identified in the dialogue texts, the (ii) location of those objects in the image, and the (iii) predicted next action.

To support development of such systems, we curated the J-ORA dataset, which provides annotations for object information in \textit{real-world environments} from the robot's egocentric perspective.
Moreover, while creating J-ORA, we introduced a standard template to annotate objects inside a dialogue scene with their attributes.  

To create J-ORA, we leverage existing \textit{dialogue text-and-image pairs} from the J-CRe3 \cite{ueda-etal-2024-j} dataset that describe true interactions between two human interlocutors to mimic a human-robot interaction setting. 
Due to the egocentric nature of the video recordings from the dialogue interactions and the consequent \textit{image sequences and audio transcripts} which are extracted from these videos, J-CRe3 is suitable for developing our perception dataset, from the robot's \textit{first-person viewpoint}.

We extend J-CRe3 by writing rich descriptions of the objects inside the images for each \textit{dialogue text-and-image pair}.
J-CRe3 contains 11,024 seconds of video (i.e.,  3 hours 3 minutes 44 seconds) of multiple dialogue scenarios. 
The following subsections outline J-CRe3, the importance of object attributes, our attribute annotation method, and statistics of our J-ORA dataset.
\subsection{J-CRe3 Dataset}
The Japanese Conversation dataset for Real-world Reference Resolution (J-CRe3) comprises egocentric video and dialogue audio recordings of real-world conversations between two people assuming the roles of a \textit{master} and an \textit{assistant robot} at home. Three laboratory environments were prepared during data collection: a kitchen, dining room, and living room. 
They initially gathered data related to 180 \textit{dialogue scenarios} from 101 workers via crowdsourcing, and the number of utterances per scenario was limited to 10 to 16. However, after filtering out the scenarios that did not meet their requirements, 142 scenarios remained and were used to construct J-CRe3. 
After that, they recruited and paired five workers to collect (i) egocentric video recordings by equipping the actor playing the robot role with a head-mounted RGB camera and third-person videos by installing four RGB cameras in four corners of the laboratory, (ii) audio recordings by equipping the actors with close-talking microphones. 
We leverage a collection of 142 dialogue scenarios from J-CRe3 comprising (i) the image sequences extracted from egocentric videos and  (ii) audio transcriptions from time-stamped utterances to construct a new dataset augmented with \textit{object attribute} information. 
\subsection{Object Attributes}
\label{subsec:ObjectAttributes}
It is crucial to introduce object descriptions to distinguish between \textit{numerous but similar objects} occurring in the same visual scene, 
for example, when three \textit{cola bottles} of different colors and sizes are placed on the same table. 
Therefore, we created a new dataset by (i) introducing a standard template developed through an iterative process and (ii) leveraging the template to collect \textit{attributes} for objects. 
Specifically, we describe objects using these features: \textit{category, color, shape, size, material, surface texture, position, state, functionality, brand, interactivity, and proximity to the person}, see Figure \ref{fig:PaperMethodObjAttr}. 
Furthermore, \textit{semantic concepts} are essential, e.g., books, notebooks, magazines, diaries are all ``books.'' Our attribute annotations distinguish semantically similar yet physically different objects.
\subsection{Changes in the Visual Scene during Dialogue}
While developing J-ORA, we deliberate on annotating object descriptions that emphasize the spatial and temporal changes occurring in the dialogue scene at different timestamps, such as; 
\textbf{1.} objects appear, disappear, and re-appear, 
\textbf{2.} the object's state changes, e.g., the bag is closed, bag is fully open, bag is partially open, bag is closed again, 
\textbf{3.} human pose and gaze change throughout the dialogue, 
\textbf{4.} objects appear inside other objects, e.g., documents placed inside a bag. 
(All four changes are shown in Figure \ref{fig:PaperObjectChanges}.)
Such annotations are vital to training robots to track changes in the scene. 
\subsection{Object Attribute Annotation}
We leverage the J-CRe3 dataset and add \textit{object attributes}, as shown in Figure \ref{fig:PaperMethodObjAttr}.  
\textit{First,} we prompt GPT-4o via API to describe objects appearing inside the image. 
Specifically, we utilize the \textit{attribute template} introduced in \cref{subsec:ObjectAttributes}, and the \textit{prompt} needed to automate the generation of object descriptions based on the template. 
The inputs to GPT-4o include the prompt, attribute template, and images. 
GPT-4o writes detailed descriptions of objects appearing in each image while keeping track of changes occurring at the different stages in the image sequence. 
\textit{Second,} we manually examine object descriptions in all dialogue scenarios and correct all the descriptions of objects that are not accurate.  
Among the 142 dialogue scenarios, there were seven scenarios where we revised the annotations, resulting in an accuracy rate of 95\%.
Figure \ref{fig:PaperMethodObjAttr} illustrates annotations of object attributes in our dataset. We have shown attribute information for \textit{backpack, and laptop}. The attribute information is described in Japanese.
During annotation, GPT-4o received the following instructions via API.
\begin{tcolorbox}[colback=gray!10, colframe=blue!75!black, width=0.995\linewidth, boxrule=0.15mm, sharp corners]
\tiny
Describe the attributes of the objects in the image.
Your descriptions must be as accurate as possible based on what you actually see in the image and contain at least 10 objects. 
Pay close attention to what the person in the image is doing, and describe the location of objects in relation to the person and other nearby objects. 
All objects and their attributes must be displayed in JSON format. Use this list of attributes for each object: 
{\color{red} category, color, shape, size, material, surface texture, position, state, functionality, brand, interactivity, and proximity to the person}.
{\color{blue} Write the attribute information in Japanese.}
\end{tcolorbox}
\subsection{Data Statistics}
The Table \ref{Table:DataStatistics} summarizes the characteristics of the proposed J-ORA dataset.
\begin{table}[t!]
    \centering
    \footnotesize
    \caption{A summary of the J-ORA dataset.}
    \resizebox{0.75\columnwidth}{!}{
    \begin{tabular}{ll}
    \toprule
    \textbf{Entry} & \textbf{Number} \\
    \hline
    \# Hours &   3 hrs 3 min 44 sec \\
    \# Unique dialogues &  93 \\
    \# Total dialogues &  142 \\
    \# Utterances &  2,131 \\
    \# Sentences &  2,652 \\
    \# Average turns per dialogue & 15 \\ 
    \# Average duration per turn & 77 sec \\
    \# Total turns & 2,131 \\
    \# Image-Dialogue pairs &   142 \\
    \# Unique object classes &  160 \\
    \# Object attribute annotations &  1,817 \\
    \bottomrule
    \end{tabular}
    }
    \label{Table:DataStatistics}
\end{table}
\section{EXPERIMENTS}
\begin{table*}[t!]
    \centering
    \caption{The Accuracy across all VLMs under the \colorbox{violet!20}{zero-shot} setting (baseline).}
    \resizebox{0.65\textwidth}{!}{
    \begin{tabular}{r||c|>{\columncolor[gray]{0.9}}c|c||c|c|>{\columncolor[gray]{0.9}}c||c}
    \toprule
    \textbf{Model} & Claude 3.5  & GPT-4o & Gemini 1.5  & LlaVa-v1.6-& LLaVA-v1.6- & Qwen2-VL 7B & EvoVLM-JP  \\
     & Sonnet &  & Pro & Vicuna-13B  &  Mistral-7B & Instruct & -v1-7B  \\
    \textbf{Perception Task} &  & &  &  & & & \\
    \hline
        Object Identification  &  93.3 & 97.6 & 88.4 & 62.9 & 32.9 & 59.9 & 64.6 \\ 
        Reference Resolution       &  80.5 & 82.5 & 82.5 & 25.8 & 10.6 & 48.8 & 52.2 \\
        Action Prediction      &  92.0 & 92.6 & 86.5 & 42.7 & 43.7 & 74.6 & 27.6 \\
        \hline
        \rowcolor{blue!15} \textbf{Model average} & 88.6 & \textbf{90.9} & 85.8 & 43.8 & 29.1 & 61.1 & 48.1 \\
    \bottomrule
    \end{tabular}
    }
    \label{tab:Zeroshot_accuracy}
\end{table*}
In this section, we leverage the abilities of VLMs and explore their effectiveness using the J-ORA dataset. 
We assess each VLM's abilities to comprehend complex and dynamically changing visual scenes and to perform three perception tasks end-to-end, under \textit{zero-shot} and \textit{fine-tuning} settings.
Due to the high cost of developing large-scale attribute-annotated Japanese datasets, we leverage an existing Japanese VQA dataset and re-format it to suit the format needed in fine-tuning VLMs for the three perception tasks.
\subsection{Experimental Settings}
\label{subsec:ModelSelectionEvaluation}
\textbf{Model Selection.} 
We chose popular VLMs, which have shown competitive results in recent studies, from three categories: \textbf{(i)} Proprietary i.e., \textit{claude-3-5-sonnet-20240620, gemini-1.5-pro, gpt-4o}. 
\textbf{(ii)} General open-source, i.e., \textit{llava-v1.6-mistral-7b-hf, llava-v1.6-vicuna-13b-hf, Qwen2-VL-7B-Instruct}.
\textbf{(iii)} Japanese open-source, i.e., \textit{EvoVLM-JP-v1-7B, japanese-stable-vlm, bilingual-gpt-neox-4b-minigpt4}.
\\ \textbf{Evaluation Metrics.}
Our evaluation methodology employs targeted task-specific prompts to extract objects, their spatial positions, and predicted actions. 
This methodological adaptation addresses our end-to-end perception requirements, by providing the \textit{accuracy} for each task in the end-to-end system. 
For quantitative assessment of selected VLMs, we calculate the scores using an LLM. 
In this study, we employ GPT-4o's API as the \textit{LLM-judge}.  
To assign scores given a \textit{dialogue image-and-text pair}, inputs to LLM-judge include five entries: \textit{the image, dialogue texts, prompts related to three perception tasks,  responses for each task, and the prompt describing the evaluation criteria}. 
The LLM judge outputs a score for each task on a scale of one to ten. 
The scores for all tasks across all image-text pairs and VLMs are converted into percentages, obtaining \textit{accuracy}.
\begin{figure*}[t!]
\begin{minipage}[t]{0.6\textwidth}
   \vspace{0pt}  
   \begin{table}[H]
       \caption{ \colorbox{green!20}{Fine-tuned} open VLM Accuracy, \textit{without attributes}.}
       \resizebox{\textwidth}{!}{
       \begin{tabular}{r||c|c|>{\columncolor[gray]{0.9}}c||c|c}
       \toprule
       \textbf{Model} & LlaVa-v1.6-& LLaVA-v1.6- & Qwen2-VL 7B & EvoVLM-JP & Japanese- \\
                      & Vicuna-13B  &  Mistral-7B & Instruct & -v1-7B & stable-vlm 7B \\
       \textbf{Perception Task} &  & &  &  &  \\
       \hline
           Object Identification   & 64.1 & 38.1 & 54.4 & 64.7 & 21.4 \\ 
           Reference Resolution        & 48.9  & 17.2 & 57.4 & 46.7 & 18.3 \\
           Action Prediction       & 50.9 & 39.4  & 57.3 & 50.5 & 19.1 \\
           \hline
          \rowcolor{blue!15}  \textbf{Model average}  & 54.6 & 31.6 & \textbf{56.3} & 53.9 & 19.6 \\
          \hline 
          $\Delta$Accuracy (vs. Baseline) & \textcolor{blue}{+10.8} & \textcolor{blue}{+2.5} & \textcolor{red}{-4.8} & \textcolor{blue}{+5.8} & \textcolor{blue}{+19.6} \\
       \bottomrule
       \end{tabular}
       }
       \label{tab:FinetuneAccuracyWattributes}
   \end{table}
   \vspace{1em}
   \begin{table}[H]
       \caption{ \colorbox{green!20}{Fine-tuned} open VLM Accuracy, \textit{with attributes}.}
       \resizebox{\textwidth}{!}{
       \begin{tabular}{r||c|c|>{\columncolor[gray]{0.9}}c||c|c}
       \toprule
       \textbf{Model} & LlaVa-v1.6-& LLaVA-v1.6- & Qwen2-VL 7B & EvoVLM-JP & Japanese- \\
                      & Vicuna-13B  &  Mistral-7B & Instruct & -v1-7B & stable-vlm 7B \\
       \textbf{Perception Task} &  & &  &  &  \\
       \hline
           Object Identification   & 67.0 & 45.8 & 74.3 & 66.4 & 38.2 \\ 
            Reference Resolution        & 46.5 & 28.9 & 67.6 & 55.2 & 29.5 \\
           Action Prediction       & 55.3 & 49.1 & 71.8 & 48.1 & 41.3 \\
           \hline
           \rowcolor{blue!15}  \textbf{Model average}  & 56.2 & 41.3 & \textbf{71.2} & 56.6 & 36.3 \\
           
          \hline 
          $\Delta$Accuracy (vs. Baseline) & \textcolor{blue}{+12.4} & \textcolor{blue}{+12.2} & \textcolor{blue}{+10.1} & \textcolor{blue}{+8.5} & \textcolor{blue}{+36.3} \\
       \bottomrule
       \end{tabular}
       }
       \label{tab:FinetuneAccuracyNoattributes}
   \end{table}
\end{minipage}%
\hfill
\begin{minipage}[t]{0.35\textwidth}
   \vspace{0pt}  
   \includegraphics[width=\textwidth]{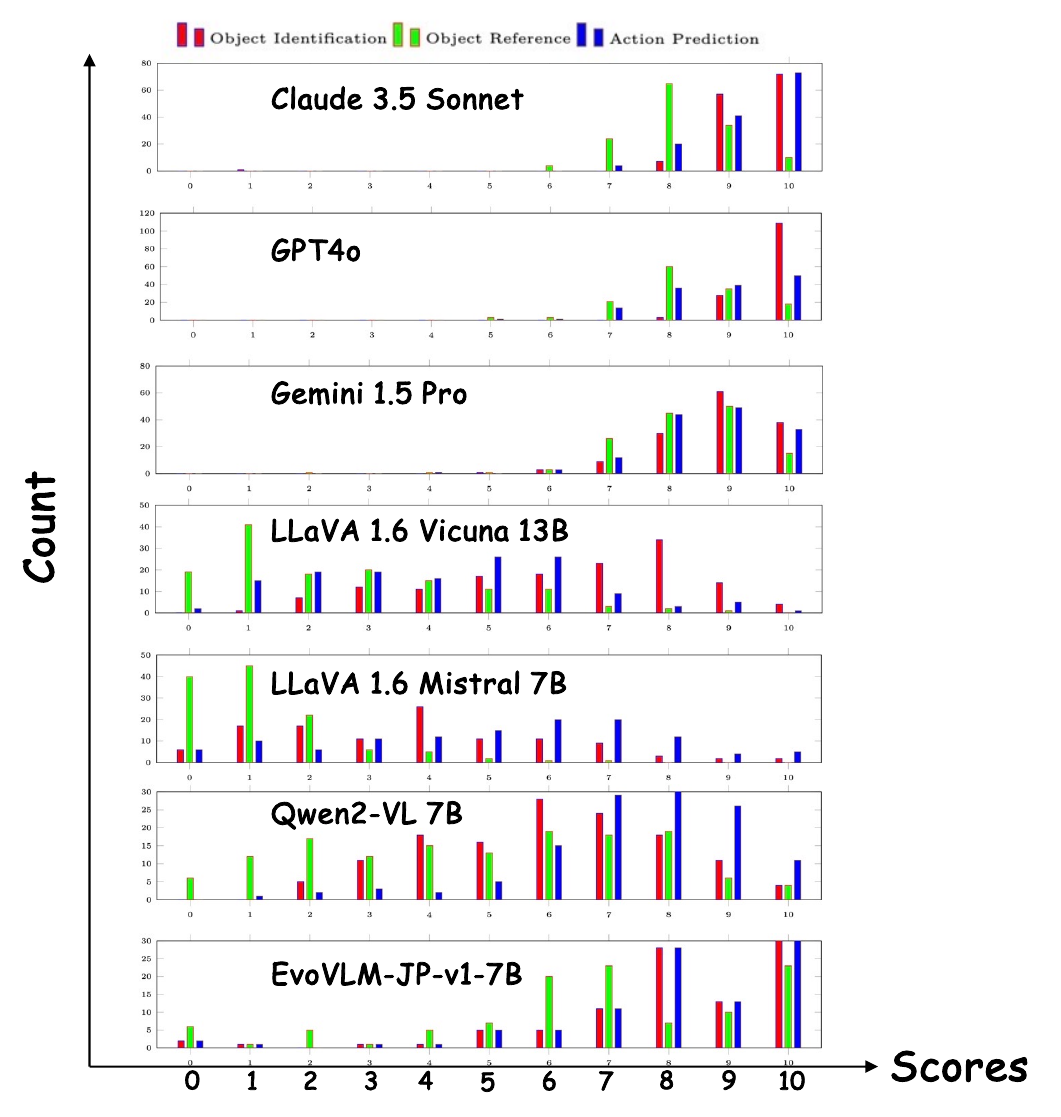}
   \caption{VLM Accuracy distributions under zero-shot settings. }
   \label{fig:ZeroshotScoreDistribution}
\end{minipage}
\end{figure*}
\\ \textbf{Zero-shot Settings.} We leverage in-context learning \cite{NEURIPS2020_1457c0d6} and write \textit{prompts} needed to elicit textual \textit{responses} from VLMs. 
The same prompts are applied to all VLMs.
\\ \textbf{Fine-tuning Settings.} 
We fine-tuned all the open-source VLMs using the re-formatted \textit{Japanese Visual Genome VQA} dataset, which contains 99K images and 793K QA pairs in Japanese. 
We used the entire data as the \textit{training set} and evaluate using our J-ORA dataset. 
We used three different seeds and fine-tuned for ten epochs using two NVIDIA A100 GPUs. 
\subsection{Results}
\textbf{Zero-shot.} Table \ref{tab:Zeroshot_accuracy} (the baseline) indicates that GPT-4o achieves the best performance with an average score of 90.9\% across three tasks. 
Other proprietary VLMs are on par with GPT-4o, and the average scores are 88.6\% and 85.8\% for Claude 3.5 Sonnet and Gemini 1.5 Pro, respectively. 
However, there is a performance gap between these VLMs and open-source counterparts. 
The best-performing open VLM is Qwen2VL-7B, with accuracy 61.1\%. The difference in accuracy between GPT-4o and Qwen2VL-7B is 29.8\%. 
Among Japanese-specific VLMs, EvoVLM 7B is on par with LlaVa 1.6 Vicuna 13B, while Japanese StableVLM 7B and Bilingual-gpt-neox-4b-minigpt4 only generated unintelligible responses, not sufficient to accomplish our task. 
Moreover, Bilingual-gpt-neox-4b-minigpt4  did not show improvement even after fine-tuning, hence it's not included in our results. These observations highlight the difficulty of our end-to-end task. 
Figure \ref{fig:ZeroshotScoreDistribution} shows the distribution of scores per perception task, and proprietary VLMs exhibited better perception abilities than open-source VLMs because the scores lie in the >80\% range for each task. 
\\ \textbf{Fine-tuning.} 
We fine-tuned all the open-source VLMs. 
As shown in Table \ref{tab:FinetuneAccuracyWattributes}, all VLMs benefited from fine-tuning with attribute data. However, Japanese-StableVLM-7B benefited most because, after fine-tuning, this VLM generated responses that were aligned with our perception task.  
Furthermore, VLMs performed better when finetuned with attributes than with no attributes. See Tables \ref{tab:FinetuneAccuracyWattributes}, \ref{tab:FinetuneAccuracyNoattributes}.
\subsection{Errors and Results Analyses} 
Overall, under next-action prediction, VLMs provided logical and well-reasoned predictions for the next action based on the dialogue, the objects mentioned, and the visual information from the image. Moreover, the proprietary VLMs did well in performing the end-to-end perception of the dialogue and visual information, and the average accuracy is more than 85\% (see Table \ref{tab:Zeroshot_accuracy}). Lastly, these VLMs exhibited a strong understanding of Japanese dialogue texts.
Conversely, we also noticed several errors in the VLM responses, and we discuss those errors below.
\\ \textbf{Proprietary VLMs.} Whereas these VLMs accurately interpreted both \textit{dialogue texts} and \textit{images}, there were occasional errors, such as (i) Incorrect extraction of objects from dialogue texts. 
Under object identification (for example, in ID:006), Gemini 1.5 Pro identified these items `side,' `work,' `confirmation,' `all-nighter,' and `sleep' as objects, but these do not qualify as \textit{physical objects} occurring in a visual scene.
(ii) Under reference resolution, VLM responses provided helpful information about the locations of the identifiable objects in the image. However, imprecise responses were observed. For example, in ID:112, Claude 3.5 Sonnet could not explicitly state that the bag and desk are not visible and that the book being held is the open document/book visible in the foreground.
\\ \textbf{Open-source VLMs.} Unlike the proprietary counterparts, this category of VLMs exhibited many weaknesses, including:
(i) Responses only repeated the instruction instead of providing a predicted interaction based on the dialogue and objects identified (for example, in ID:001). 
(ii) Response is unintelligible and irrelevant to the perception task (for example, in ID:003). 
(iii)  In next-action prediction, VLM could not provide any prediction for the next course of the interaction based on the task description; instead, it repeated an unrelated phrase about answering the question beneath a given image (for example, in ID:004). 
(iv) Confusing responses due to unclear actions and random numbers presented without context (for example, in ID:018). 
\begin{table*}[t]
    \centering
    \caption{Affordance of VLMs for major objects in the dialogues under \colorbox{violet!20}{zero-shot} setting (baseline). }
    \resizebox{0.65\textwidth}{!}{
    \begin{tabular}{r||>{\columncolor[gray]{0.9}}c|c|c||c|c|c||c}
    \toprule
    \textbf{Model} & Claude 3.5  & GPT-4o & Gemini 1.5  & LlaVa-v1.6-& LLaVA-v1.6- & Qwen2-VL 7B & EvoVLM-JP  \\
     & Sonnet &  & Pro & Vicuna-13B  &  Mistral-7B & Instruct & -v1-7B \\
    \textbf{Object} &   &  &   &  &   &   &  \\
    \hline
        \begin{CJK*}{UTF8}{gbsn}かばん\end{CJK*} (Bag)              &  67.2  & 62.6  & 53.5  & 42.7  & 3.7 &  39.0  & 15.4 \\ 
        \begin{CJK*}{UTF8}{gbsn}本\end{CJK*} (Book)                &  77.0  & 70.5  & 59.6  & 67.7  & 13.4 & 54.2 & 31.0 \\
        \begin{CJK*}{UTF8}{gbsn}専門書\end{CJK*} (Technical Book)   &  78.2  & 25.4  & 38.1  & 64.2  & 7.4  & 35.1  & 30.9  \\
        \begin{CJK*}{UTF8}{gbsn}コーヒー0\end{CJK*} (coffee 0)      &  65.5  & 47.7  & 19.7 & 63.1  & 28.8 & 11.4  &  40.7 \\
        \begin{CJK*}{UTF8}{gbsn}コーヒー1\end{CJK*} (coffee 1)      &  64.2  & 49.2  & 16.9  & 62.8  & 21.5  & 13.7  & 36.2 \\
        \begin{CJK*}{UTF8}{gbsn}コーヒー2\end{CJK*} (coffee 2)      &  67.1  & 50.3  & 16.1  & 62.3  & 21.7  & 14.3  & 36.3  \\
        \begin{CJK*}{UTF8}{gbsn}書類0\end{CJK*} (Document 0)       &  64.2  & 42.9  & 76.7 & 64.9  & 27.6  & 24.9  & 59.0 \\
        \begin{CJK*}{UTF8}{gbsn}書類1\end{CJK*} (Document 1)       &  66.7  & 56.1  & 65.1  & 66.6  & 31.6  & 25.4  & 59.1 \\
        \begin{CJK*}{UTF8}{gbsn}書類2\end{CJK*} (Document 2)       &  60.8  & 41.7  & 60.7  & 62.7  & 30.3  & 25.0  & 55.5 \\
        \begin{CJK*}{UTF8}{gbsn}棚\end{CJK*} (Shelf)               &  74.4  & 61.8  & 57.0 & 61.9 & 7.5  & 20.5  & 38.7   \\
        \begin{CJK*}{UTF8}{gbsn}机\end{CJK*} (Table)               &  65.8  & 57.1  & 49.8  & 55.7  & 6.8  & 45.7  & 1.8 \\
        \hline
         \rowcolor{blue!15} Composite Affordance  & \textbf{68.3}  & 51.4  & 46.7  & 61.3  & 18.2  & 24.9  & 34.2 \\
    \bottomrule
    \end{tabular}
    }
    \label{tab:Zeroshot_Affordances}
\end{table*}
\begin{table*}[t!]
    \centering
    \caption{ Affordance of the major objects for \textit{open VLMs} under \colorbox{green!20}{fine-tuning} settings, with dialogue texts and attribute information in Japanese. 
    All \textit{open VLMs} except LlaVa-v1.6-Vicuna-13B benefit from this fine-tuning.}
    \resizebox{0.65\textwidth}{!}{
    \begin{tabular}{r||>{\columncolor[gray]{0.9}}c|c|c||c|c}
    \toprule
    \textbf{Model} & LlaVa-v1.6- & LLaVA-v1.6- & Qwen2-VL 7B & EvoVLM-JP & Japanese \\
                   & Vicuna-13B  &  Mistral-7B & Instruct    & -v1-7B    & -stable-vlm-7B \\
    \textbf{Object} & &  &   &   &  \\
    \hline
        \begin{CJK*}{UTF8}{gbsn}かばん\end{CJK*} (Bag)              &  55.1  & 21.8   &  42.3  & 20.6  & 7.5  \\ 
        \begin{CJK*}{UTF8}{gbsn}本\end{CJK*} (Book)                &  46.4  & 28.4  &  57.1  & 34.2  & 36.0 \\
        \begin{CJK*}{UTF8}{gbsn}専門書\end{CJK*} (Technical Book)   &  47.7  & 23.0   &  38.1  & 33.1  & 18.7 \\
        \begin{CJK*}{UTF8}{gbsn}コーヒー0\end{CJK*} (coffee 0)      &  55.5  & 28.8  &  36.1  & 45.1  & 6.6  \\
        \begin{CJK*}{UTF8}{gbsn}コーヒー1\end{CJK*} (coffee 1)      &  50.1  & 21.5  &  30.7  & 38.7  & 7.3  \\
        \begin{CJK*}{UTF8}{gbsn}コーヒー2\end{CJK*} (coffee 2)      &  48.3  & 21.7  &  35.8  & 37.5  & 6.1  \\
        \begin{CJK*}{UTF8}{gbsn}書類0\end{CJK*} (Document 0)       &  49.8  & 6.4  &  40.7  & 60.0  & 23.0 \\
        \begin{CJK*}{UTF8}{gbsn}書類1\end{CJK*} (Document 1)       &  43.3  & 5.6  &  43.4  & 56.8  & 22.3 \\
        \begin{CJK*}{UTF8}{gbsn}書類2\end{CJK*} (Document 2)       &  39.5  & 6.8  &  39.4  & 61.4  & 22.7 \\
        \begin{CJK*}{UTF8}{gbsn}棚\end{CJK*} (Shelf)              &  55.8  & 39.3   &  36.3  & 41.3   & 11.2 \\
        \begin{CJK*}{UTF8}{gbsn}机\end{CJK*} (Table)              &  43.9  &  24.0  &  47.5  & 14.2   & 10.3 \\
      \hline
      \rowcolor{blue!15} Composite Affordance score &  \textbf{48.7} & 20.7  & 40.7  & 40.3 & 13.6 \\
      \hline
      $\Delta$Affordance (vs. Baseline) & \textcolor{red}{-12.6} & \textcolor{blue}{+2.5} & \textcolor{blue}{+15.8} & \textcolor{blue}{+6.1} & \textcolor{blue}{+13.6} \\
    \bottomrule
    \end{tabular}
    }
    \label{tab:Finetune_Affordance}
\end{table*}
\begin{table*}[t!]
    \centering
    \caption{Accuracy for VLM under \colorbox{violet!20}{zero-shot} settings. We translated dialogue texts into English and omit Japanese-specific VLMs (EvoVLM-JP-v1-7B and Japanese-stable-vlm-7B) because they have limited English ability. }
    \resizebox{0.65\textwidth}{!}{
    \begin{tabular}{r||>{\columncolor[gray]{0.9}}c|c|c||c|c|>{\columncolor[gray]{0.9}}c }
    \toprule
    \textbf{Model} & Claude 3.5  & GPT-4o & Gemini 1.5  & LlaVa-v1.6-& LLaVA-v1.6- & Qwen2-VL 7B  \\
                 & Sonnet &  & Pro & Vicuna-13B  &  Mistral-7B & Instruct \\
    \textbf{Perception Task} &  & &  &  & & \\
    \hline
        Object Identification  & 93.1  & 92.5 & 91.7 & 51.9 & 42.6 & 68.1 \\
         Reference Resolution       & 82.3  & 87.4 & 89.3 & 53.3 & 21.2 &  59.6 \\
        Action Prediction      & 92.8  & 86.4 & 83.2 & 64.4 & 45.7 & 70.7  \\
        \hline
        \rowcolor{blue!15} Model average  & \textbf{89.4}  & 88.8 & 87.1 & 56.5 & 36.5 & 66.1 \\ 
        \hline
        $\Delta$Accuracy (vs. Baseline) & \textcolor{blue}{+0.8} & \textcolor{red}{-2.1} & \textcolor{blue}{+1.3} & \textcolor{blue}{+12.7} & \textcolor{blue}{+7.4} & \textcolor{blue}{+5.0} \\
    \bottomrule
    \end{tabular}
    }
    \label{tab:Zeroshot_accuracyEnglish}
\end{table*}
\subsection{Object Affordance} 
\label{subsec:ObjectAffordance}
We investigated the VLM abilities to recognize potential actions that can be performed with an object, i.e., the \textit{affordance}. 
Whereas other metrics such as accuracy and f1-score measure the overall performance of the VLMs, we further investigated the VLM abilities to comprehend the \textit{role} of each major object in the dialogue/scene. 
Our focus is to demystify the VLM's ability to locate an object within the dialogue, ground that object to its correct position in the visual scene, and use this information to determine what actions the robot can perform with that object.    
Therefore, we provided the affordance scores for this group of key objects: \textit{bag, book, technical book, document, shelf, and table}.
Affordance scores are shown in Table \ref{tab:Zeroshot_Affordances}, and Claude 3.5 Sonent had the highest affordance of 68.3\%. 
Our experiments reveal that high performance on perception tasks in this work does not necessarily lead to high object affordance. 
This is because not only is the object's identification and grounding in the scene vital, but it also requires a deep understanding of the range of actions that can be performed with an object. 
Although our attributes include \textit{functionality}, it is impossible to list all the possible use cases to which an object can be applied. The VLM is expected to extrapolate based on the context. 
Yet the VLMs mostly predicted the unsophisticated actions (e.g., putting and removing books on a shelf) that an object could perform, leaving out the more complex actions (e.g., use of the shelf for re-arranging books and improving the interior decor of a room). 
These findings highlight the need to expand VLMs' understanding of object affordances in a visual and interactive context. 
Furthermore, fine-tuning benefits the open VLMs as shown in Table \ref{tab:Finetune_Affordance}.
\subsection{Human Evaluation Alignment}
\label{subsec:AlignmentHumanEval}
We assessed the usefulness of GPT-4o as the main evaluator in our work. 
First, we selected 100 responses from LlaVa-v1.6-Vicuna-13B, for appraisal by human evaluators. 
Then, we recruited two human annotators to review the VLM responses and assign their scores for each response. 
By calculating the intraclass correlation coefficient, the agreement between human annotators and a GPT-4o evaluator is 85\%. 
The high level of agreement suggests GPT-4o performs comparably to human annotators in assessing the perception tasks.
\subsection{Ablation Studies}
We conducted experiments to distinguish between VLMs' language and vision abilities after translating\footnote{We leverage ChatGPT-4 to translate the dialogue texts from Japanese to English.} dialogue texts into English. 
VLMs perform better than the baseline with English dialogue texts, see Table \ref{tab:Zeroshot_accuracyEnglish}.
\section{CONCLUSIONS}
In this paper, we assume a robot that assists users in the home, and we define attribute information that contributes to the typical tasks of object detection, reference resolution, and action prediction.
We defined attribute information contributing to these tasks and constructed the J-ORA dataset by extending the human-robot dialogue dataset in our living space.
With our framework, the dataset can be constructed semi-automatically, and efficient data expansion can be expected in the future.
We evaluated the existing VLM using the constructed dataset.
We investigated the contribution of the attribute information we defined to these robot tasks toward realizing an end-to-end perception model for robots.
One future work is the size of the dataset.
Model tuning is necessary to use this attribute information appropriately, and our framework is expected to construct large-scale datasets that can respond to a variety of situations.

\bibliographystyle{IEEEtranS}
\bibliography{root}
\end{document}